\documentclass[sigconf]{acmart}
\AtBeginDocument{%
  }

\copyrightyear{2026}
\acmYear{2026}
\setcopyright{cc}
\setcctype{by}
\acmConference[SIGIR '26]{Proceedings of the 49th International ACM SIGIR Conference on Research and Development in Information Retrieval}{July 20--24, 2026}{Melbourne, VIC, Australia}
\acmBooktitle{Proceedings of the 49th International ACM SIGIR Conference on Research and Development in Information Retrieval (SIGIR '26), July 20--24, 2026, Melbourne, VIC, Australia}
\acmDOI{10.1145/3805712.3808428}
\acmISBN{979-8-4007-2599-9/2026/07}

\begin{document}

\title{Variance Reduction for Heavy-Tailed Monetization Metrics in Ranking Experiments via Post-Stratification}

\author{Neeti Pokharna}
\email{neetipokharna@sharechat.co}
\orcid{0009-0003-7089-1515}
\affiliation{%
  \institution{ShareChat}
  \city{Bengaluru}
  \state{Karnataka}
  \country{India}
}

\author{Olivier Jeunen}
\authornotemark[1]
\email{olivierjeunen@gmail.com}
\orcid{0000-0001-6256-5814}
\affiliation{%
  \institution{Aampe}
  \city{Antwerp}
  \country{Belgium}
}

\author{Yatharth Saraf}
\email{yatharth@sharechat.co}
\orcid{0009-0000-3619-2380}
\affiliation{%
  \institution{ShareChat}
  \city{London}
  \country{UK}
}

\author{Aleksei Ustimenko}
\authornote{Work done while at ShareChat}
\email{me@aleksei.uk}
\orcid{0009-0006-4942-7779}
\affiliation{%
  \institution{Simulacra Research}
  \city{London}
  \country{UK}
}

\renewcommand{\shortauthors}{Neeti Pokharna, Olivier Jeunen, Yatharth Saraf, \& Aleksei Ustimenko}

\begin{abstract}
Online evaluation of ranking and retrieval systems often relies on downstream monetization metrics such as app revenue or creator earnings. These metrics are typically heavy-tailed, with a small fraction of users dominating both mean and variance, leading to low statistical power and unreliable conclusions in A/B experiments---especially under limited traffic. 

We present a practical framework for variance reduction in online experiments by combining post-stratification with CUPED. Our approach leverages pre-experiment covariates to improve the sensitivity of monetization experiments without requiring additional traffic. Deployed at ShareChat across ranking-driven monetization experiments, the method substantially reduces variance and improves decision stability, achieving equivalent statistical confidence with ~45\% less traffic than standard metrics. We further discuss practical design choices, guardrails, and limitations, providing guidance on when post- stratification is appropriate for real-world information retrieval and Recommendation systems.
\end{abstract}




\maketitle
\section{Introduction and Motivation}

Online controlled experiments are the primary mechanism for evaluating ranking and recommendation systems in large-scale Information Retrieval (IR) platforms. Because ranking systems control exposure and discovery, they indirectly shape downstream monetization outcomes, making monetization-sensitive evaluation a core IR measurement challenge rather than purely a business metric problem. Changes to feed personalization are typically validated through A/B tests that require sufficient statistical power~\cite{Kohavi2009} and stable estimation of treatment effects. While engagement metrics are well studied in IR evaluation, many production systems now optimize for monetization objectives---including advertising revenue, marketplace liquidity, and creator earnings---which introduce distinct statistical challenges.

Unlike engagement metrics, monetization outcomes are typically driven by rare but high-impact user behavior and exhibit heavy-tailed distributions. Under realistic traffic allocations, this tail behavior dominates variance and makes normal approximations derived from large-sample theory unreliable in practice~\cite{Jeunen2025}. As a result, treatment effect estimates become unstable, reducing experimental sensitivity and leading to unreliable inference under practical traffic and duration constraints~\cite{Jeunen2024}.

While variance-reduction techniques like CUPED (Controlled Experiments Using Pre-Experiment Data)~\cite{Deng2013} improve estimator efficiency, their effectiveness is limited by the predictive strength of pre-experiment signals. 
Operationally, this forces teams to extend experiment durations or relax statistical thresholds, slowing iteration. 

Post-stratification, a classical technique from survey sampling~\cite{valliant1992,Miratrix2013}, addresses population heterogeneity by partitioning units into homogeneous strata and aggregating stratum-level estimates using population-level weights~\cite{Xie2016}. Applying it to heavy-tailed monetization outcomes in production IR systems introduces new challenges, including robust stratum design, unbiased estimation under skewed distributions, and seamless integration into existing experimentation pipelines.

In this work, we introduce a production-ready post-stratification framework tailored for IR experimentation. Deployed at ShareChat and Moj across $>40$ ranking-driven monetization experiments, our approach: (i) analyzes the statistical behavior of heavy-tailed metrics under traffic constraints; (ii) combines behavioral post-stratification with covariate adjustment; (iii) demonstrates consistent variance reduction, enabling reliable decisions and faster iteration on ranking improvements.

\section{Problem Statement \& Methodology}
\subsection{Monetization Metrics}
ShareChat's live streaming platform enables users to purchase virtual currency and gift it to content creators during live sessions. GMV (Gross Merchandise Value) represents the total recharge amount during the experiment window, serving as the platform's primary monetization metric for digital live gifting~\cite{Jeunen2024}. GMV is a downstream outcome of multiple IR components, including retrieval, ranking, diversification, and recommendation systems. Changes to ranking models directly affect exposure, which in turn shapes monetization behavior.

The unit of randomization in our online experiments is the user. Let $Y(1)$ and $Y(0)$ denote post-experiment GMV under treatment and control, respectively. Our target estimand is the population Average Treatment Effect (ATE) $\tau = \mathbb{E}[Y(1) - Y(0)]$. We seek an estimator $\hat{\tau}$ that is unbiased, has lower variance than the difference-in-means, and is robust to implementation choices and sample variations.

\subsection{Distributional Challenges in Monetization Metrics}

\begin{figure}[t]
  \centering
  \includegraphics[width=\linewidth]{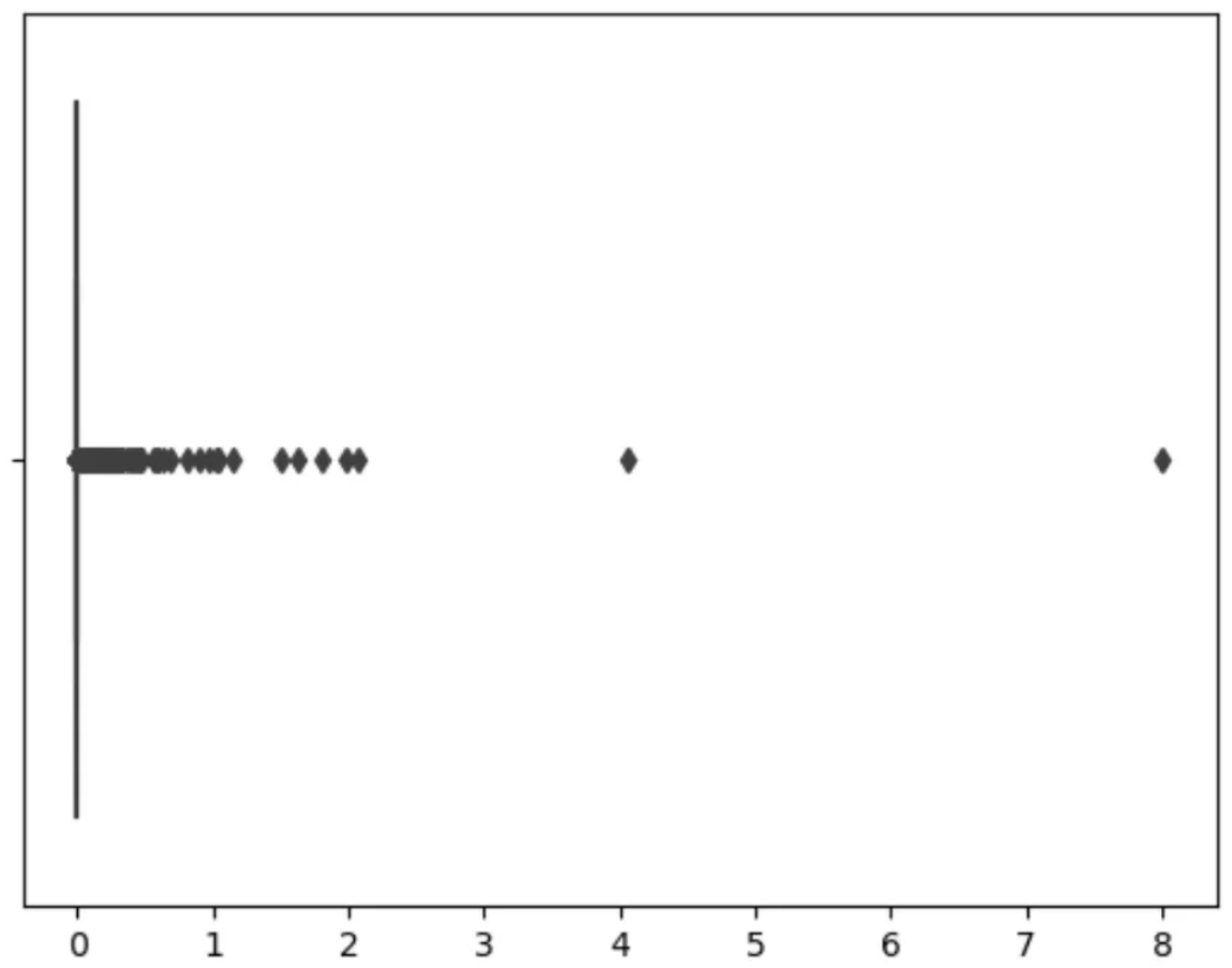}
  \caption{Distribution of Pre-experiment Raw GMV (total GMV calculated using pre-experiment data at user level). The box plot (x-axis values omitted for confidentiality) shows the median, quartiles, and outliers. The extreme outliers contribute disproportionately to variance, causing statistical instability.}
  \Description{Box plot of Raw GMV showing extreme outliers}
  \label{fig:raw-gmv}
\end{figure}

We analyzed the distribution of GMV and observed extreme skewness, consistent with heavy-tailed or power-law-like behavior commonly observed in user activity and monetization data~\cite{Clauset2009}. As shown in Figure~\ref{fig:raw-gmv}, the top 0.01\% of users dominate the variance, creating a heavy tail that defies standard approximations.

\begin{figure}[t]
  \centering
  \includegraphics[width=\linewidth]{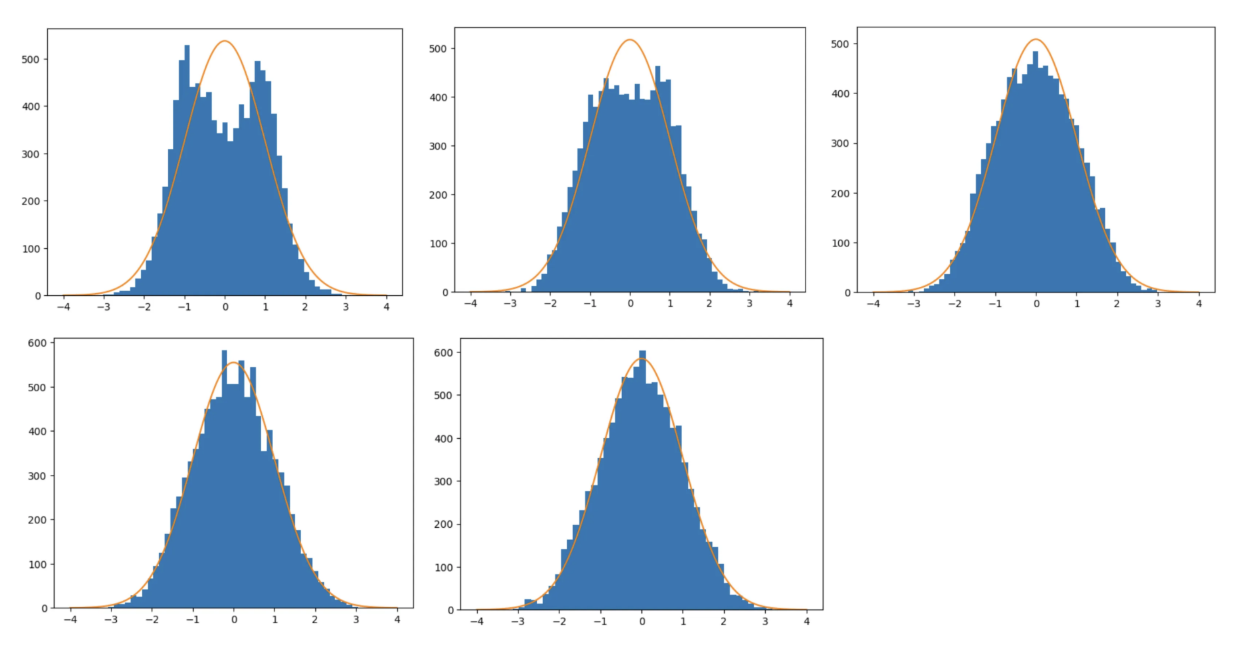}
  \caption{Empirical distribution of $z$-statistics for GMV (x-axis) from A/A tests at different traffic levels (Top: 1\%, 5\%, 10\%; Bottom: 20\%, 30\%). The y-axis represents frequency. Deviations from the standard normal (orange curve) indicate CLT failure.}
  \Description{Empirical distribution of $z$-statistics}
  \label{fig:gmv-distribution}
\end{figure}

To validate statistical assumptions, we analyzed the distribution of $z$-statistics across thousands of A/A simulation runs at varying traffic levels (1\%, 5\%, 10\%, 20\%, 30\%, 50\%). While the Central Limit Theorem (CLT) guarantees normality asymptotically, the heavy skew of GMV delays this convergence significantly. As shown in Figure~\ref{fig:gmv-distribution}, at low-to-moderate traffic (1--20\%), the empirical distribution of $z$-statistics deviates from the theoretical standard normal. Convergence begins at 20\% but remains incomplete; satisfactory adherence to the standard normal is only observed at $\geq$50\% traffic. 

Crucially, this violation manifests as conservatism in hypothesis testing: we observed False Positive Rates (FPR) of 2--4\% compared to the nominal 5\% predicted by theory. While this implies that any statistically significant result is highly trustworthy (low Type-I error), it also indicates that the test is severely underpowered. In standard A/B testing at 5--10\% traffic, the probability of detecting true effects becomes negligible without variance reduction. In practice, this implies that simply increasing experiment duration or traffic is insufficient to recover statistical reliability for heavy-tailed monetization metrics.

\subsection{Post-Stratification Framework}
Post-stratification helps control for heterogeneity in user behavior, especially in the distribution's tail. The idea is to partition users into discrete strata based on observed characteristics---such as past revenue activity---and compute experiment effects within each stratum.
These stratum-level results are then aggregated using population-level weights based on the proportion of each stratum in the full platform population, not just experiment participants. This ensures external validity and prevents bias from treatment-control imbalances within strata.
This method reduces residual variance substantially without introducing bias~\cite{Miratrix2013}, thereby improving statistical power. Additionally, post-stratification improves interpretability, as experiment owners can observe whether treatment effects are concentrated in tail users or evenly distributed across strata.

\textbf{Critical distinction:} Strata are defined based on pre-period GMV, ensuring that the treatment cannot affect stratum membership. 

Let $S = \{1, 2, \ldots, k\}$ denote strata (e.g., tail vs. non-tail users). Our post-stratified estimator is:
\[
\hat{\tau}_{PS} = \sum_{s \in S} w_s \cdot \hat{\tau}_s^{CUPED}
\]
where $w_s = N_s / N$ is the stratum's population proportion, $N_s$ is the number of users in stratum $s$ across the full platform, and $\hat{\tau}_s^{CUPED}$ is the CUPED-adjusted treatment effect in stratum $s$. 

Our approach follows principles from regression adjustment in randomized experiments, which can improve estimator efficiency without requiring correct model specification~\cite{Lin2013}.

\textbf{Unbiasedness:}
\[
\mathbb{E}[\hat{\tau}_{PS}]
    = \sum_{s \in S} w_s \, \mathbb{E}[\hat{\tau}_s^{\text{CUPED}}]
    = \sum_{s \in S} w_s \, \tau_s
    = \mathbb{E}_{\mathcal{P}}[\tau]
       = \tau
\]
\textit{as CUPED is unbiased \cite{Deng2013} and by the law of total expectation}. This result follows from Theorem~1 of \citeauthor{Miratrix2013}~\cite{Miratrix2013}.
The unbiasedness requires that strata are defined using only pre-treatment information. Since we construct strata from pre-period GMV, stratum membership is unaffected by the treatment assignment, ensuring the conditions for unbiased post-stratified estimation.
\begin{enumerate}
\item \textbf{Stratification:} Isolating high-variance users (tail) into a small-weight stratum reduces their contribution to overall variance by the squared weight factor. For a tail stratum with population weight $w \approx 0.0001$, its contribution to post-stratified variance is $w^2 \cdot \sigma^2_{\text{tail}} / n_{\text{tail}}$. Even if $\sigma^2_{\text{tail}}$ is enormous, multiplication by $\sim$$10^{-8}$ effectively suppresses it.
\item \textbf{CUPED within strata:} Pre-period covariates remove predictable variation within each homogeneous stratum
\end{enumerate}

\section{Experimental Validation and Results}
\label{sec:experimental-validation}
We validated our methodology on over 40 production A/B tests affecting Live streaming ranking and recommendation systems, each with more than 1M users. Results were consistent across both platforms - ShareChat and Moj — demonstrating robustness across product surfaces.

Our workflow: 
\begin{enumerate}
\item Collect raw and pre-experiment GMV per user.
\item Winsorize or remove values beyond the 99.999th percentile (typically $<5$ users). As shown in Figure~\ref{fig:raw-gmv}, this mitigates the impact of extreme power users (e.g., the 2 outliers visible here) that otherwise dominate variance, without biasing the vast majority of the population.
\item Assign users to behavioral strata (tail vs. non-tail vs new users/non-spenders) using 30-day historical GMV. Stratification thresholds are computed from the pre-period population and frozen before any experiment outcome data is observed.
\item Compute CUPED-adjusted GMV within each stratum:
\begin{align*}
  \text{GMV}_{\text{CUPED}} &= \text{GMV} \\
  & - \frac{\text{cov}(\text{GMV}, \text{CentralizedPreGMV})}{\sigma^2_{\text{CentralizedPreGMV}}} \cdot (\text{CentralizedPreGMV})
\end{align*}

  where: 
  \[
    \text{CentralizedPreGMV} = \text{PreGMV} - \mu_{\text{PreGMV}}
  \]
\item Compute post-stratified mean and variance of $\text{GMV}_{\text{CUPED}}$ using full-traffic weights:
  \[
    \mu_{\text{treatment}} = \sum_{s \in S} w_s \cdot \mu_s \qquad 
    \sigma^2_{\text{treatment}} = \sum_{s \in S} w_s^2 \cdot \frac{\sigma^2_s}{n_{T,s}}
  \]
  Similarly, calculate the same metric for the control group.
\item Calculate the $z$-statistic and $p$-value to assess statistical significance.
\[
    \text{z-statistic}=\frac{\mu_{\text{treatment}} - \mu_{\text{control}}}{\sqrt{{\sigma^2_{\text{treatment}}}+{\sigma^2_{\text{control}}}}}
  \]
\end{enumerate}

\begin{table}[h]
  \setlength{\tabcolsep}{6pt} 
  \caption{Comparison of Variance Reduction \& Type-I Error across GMV metric variants}
  \label{tab:results}
  \centering
  \small
  \begin{tabular}{lrrr} 
    \toprule
    \textbf{Approach} & \textbf{Var. Reduction} & \textbf{Med. rel. Z} & \textbf{Type-I Error} \\
    \midrule
    Raw GMV & - & - & 1.0\% \\
    CUPED-adjusted GMV & 47.62\% & 1.10 & 2.6\% \\
    Post-strat (winsorized) & 99.3\% & 1.35 & 6.1\% \\
    Post-strat (raw tails) & 99.7\% & 1.36 & 6.1\% \\
    \bottomrule
  \end{tabular}
\end{table}

Compared to raw GMV, CUPED reduces variance by $\sim$48\%, while post-stratification with CUPED achieved $>99\%$ reduction in this case. This extreme reduction is driven by down-weighting a small fraction of users (<0.01\%) who dominate variance, yielding \textbf{improved sensitivity} (Table~\ref{tab:results}) and enabling detection of subtle treatment effects previously masked by noise. The negligible difference (0.4\%) between two post-stratification approaches confirms that isolating tail users—rather than specific outlier treatment—is the primary driver. 

In practical terms, the variance reduction translates to substantially lower minimum detectable effects (MDE): at 10\% traffic allocation, MDE drops from $\sim$136\% of the mean (raw GMV) to $\sim$10\% (post-stratified), enabling detection of realistic treatment effects that were previously infeasible.

Beyond variance reduction, we evaluate decision quality by measuring \textbf{time-to-decision reduction} under fixed traffic, \textbf{consistency} with long-running baselines, and \textbf{sensitivity} to tail events. Across experiments, post-stratification enables reliable decisions using 40--50\% less traffic, while maintaining agreement with long-horizon outcomes.

\textbf{Interpretation of Median Relative Z-score:}
A median relative $z$-score of 1.36 implies that we need $1.36^2=1.85$ times fewer data points to achieve the same level of confidence as the original metric $(1 - \frac{1}{1.85} \approx 45\%)$~\cite{Baweja2024}. For comparison, the traffic reduction enabled by CUPED-adjusted GMV is $\approx17\%$.

The corresponding $z$-score increase is more modest than the variance reduction because $z$-scores scale with the inverse square root of variance, and down-weighting tail users inherently shrinks the mean difference (numerator) alongside the noise. Some experiments have genuine effects concentrated in tail users, which are appropriately down-weighted.

\textbf{Type-I Error Analysis:} 
The two Type-I error figures in Table~\ref{tab:results} and Section~\ref{sec:robustness} measure different things and are not in contradiction. The 6.1\% figure in Table~\ref{tab:results} is an \emph{empirical} rate measured across our 40+ production A/B tests, which involve real heterogeneous user populations, live traffic, and potential minor experiment contamination. The 0\% figure reported in Section~\ref{sec:robustness} is a \emph{simulation-based} rate obtained by repeatedly splitting a static pre-experiment snapshot into random A/A halves under controlled conditions. The simulation environment removes real-world noise sources such as novelty effects, carryover, and randomization imbalance so it naturally yields lower false positive rates. Both figures are consistent with post-stratification functioning correctly: the simulation confirms the estimator is unbiased under ideal conditions, while the production figure reflects slightly elevated but acceptable Type-I error ($\sim$6.1\% vs. nominal 5\%) due to finite-sample stratum variance estimation.
 
Post-stratification yields this slightly elevated empirical Type-I error rate due to finite-sample stratum variance estimation and potential effect heterogeneity across strata. However, raw and CUPED-only metrics are severely underpowered (1--2.6\% false positive rates). A Type-I error rate far below nominal does not indicate conservatism---it indicates a lack of statistical power and is essentially uninformative. We accept this slight elevation because the massive reduction in Type-II error (False Negative Rate) allows us to detect valuable ranking improvements that were previously statistically invisible. Consequently, we treat the post-stratified metric as sufficient for \textit{low-risk ranking changes}, while \textit{high-risk launches} require confirmation over longer horizons.

\textbf{Estimand Interpretation:}
A key question is whether the post-stratified ATE is the right estimand when tail users contribute the majority of revenue. Post-stratification down-weights tail users (weight $w \approx 0.0001$) to suppress variance, which means the post-stratified estimate is sensitive to ranking changes that affect the broad non-tail population, but is relatively insensitive to effects concentrated exclusively in whales. This is a deliberate design choice with concrete business implications: if a ranking change primarily benefits non-tail users while being neutral or slightly negative for high-value users, post-stratification will correctly detect it as positive, while raw GMV may fail to reach significance at all. Conversely, if the primary business goal is to evaluate effects on whale users specifically---for example, a VIP program or a monetization feature targeting heavy spenders---the post-stratified metric is \emph{not} the right tool, and we explicitly recommend against it in that case (see Section~\ref{sec:deployment}). For general ranking experiments where the goal is to improve the overall user experience and monetization broadly, the post-stratified ATE is the appropriate estimand. We surface both raw and post-stratified metrics to experiment owners so that directional disagreement between the two (a proxy for tail-concentrated effects) is always visible.

\textbf{Winsorization Sensitivity:}
The 99.999th percentile winsorization threshold (affecting typically $<5$ users per experiment) was chosen conservatively to remove only unambiguous extreme outliers. To assess sensitivity, we varied the threshold from the 99.9th to 99.999th percentile. Variance reduction at the 99.9th percentile was 95.2\% (Table~\ref{tab:threshold-sensitivity}), confirming that winsorization at this level contributes modestly relative to stratification itself. The negligible difference between Post-strat (with winsorization) and Post-strat (raw tails) in Table~\ref{tab:results} (99.3\% vs. 99.7\%) further confirms that winsorization has a minor incremental effect: the dominant variance driver is stratum weight suppression, not outlier removal. We retain the 99.999th percentile threshold as a lightweight safeguard against data pipeline errors or anomalous single-user events.
 
To validate that detected effects are genuine and not tail noise, we examined temporal consistency: effects detected in week 1 maintained 91\% directional agreement in week 2 across all 40+ experiment-variant pairs.

Following validation, we deployed post-stratified GMV as the default metric for all monetization experiments at ShareChat. We also extended the framework to gaming revenue and coins earned metrics increasing the number of statistically conclusive experiments. Across all launches evaluated using post-stratified GMV, we observed no instances where detected effects failed to reproduce at higher traffic or required rollback. Average time-to-decision was reduced by $\sim 50\%$.

\section{Robustness Checks and Failure Modes}
\label{sec:robustness}

\begin{table}[h]
\setlength{\tabcolsep}{3pt}
\caption{Variance Reduction by Tail Threshold}
\label{tab:threshold-sensitivity}
\centering
\begin{tabular}{lcc}
\toprule
\textbf{Threshold} & \textbf{Var. Reduction} & \textbf{Median $n_{tail}$} \\
\midrule
99.9\%ile & 95.2\% & 500 \\
99.95\%ile & 97.8\% & 250 \\
99.99\%ile & 99.3\% & 100 \\
99.995\%ile & 99.7\% & 50 \\
\bottomrule
\end{tabular}
\end{table}
\textbf {Sensitivity to Stratification Thresholds:}
We evaluated sensitivity to stratification thresholds by varying the tail cutoff between the 99.9th and 99.995th percentiles (Table~\ref{tab:threshold-sensitivity}). 
Variance reduction remained above 95\% across all thresholds, with optimal performance at 99.99th percentile. Statistical conclusions (ship/no-ship decisions) were identical across thresholds in 95\% of experiments, confirming robustness to threshold choice. 

\textbf{Stratum Design Exploration:}
Beyond percentile-based stratification on pre-period GMV, we evaluated multi-dimensional stratification using behavioral features including agency creator status, quality DAU flags, new recharger indicators, and combinations thereof via brute-force search. Feature-based strata achieved 25--35\% variance reduction, while k-means clustering on user features actually increased variance. Simple percentile-based GMV stratification consistently dominated ($>$60\% variance reduction before CUPED), confirming that outcome-based stratification is substantially more effective than behavioral segmentation for heavy-tailed metrics.

\textbf {A/A Validation Under Post-Stratification:} 
We conducted A/A experiments using the post-stratified estimator. While empirical Type-I error increased relative to raw GMV, it remained stable across independent A/A runs and did not exhibit systematic bias toward positive effects. False positives were not concentrated in any single stratum. In simulation-based sensitivity analysis, post-stratified metrics yielded 0\% false positives across traffic levels from 5\% to 50\%, compared to 1--7\% for raw GMV. As discussed in Section~\ref{sec:experimental-validation}, the difference between this simulation result and the 6.1\% empirical figure reflects the gap between controlled conditions and live production traffic.

\textbf{Failure Modes:}
In 1 of 40+ experiments, post-stratification increased variance rather than reducing it, attributable to a treatment effect concentrated entirely in the tail stratum combined with stratum-level sample imbalance. This reinforces the guidance that the method should not be used when tail user behavior is the primary decision objective.

\textbf {Why Not Alternative Approaches?}
We evaluated alternative approaches including log-transformed metrics, trimmed means, and non-parametric tests. Log-transformation ($\text{LogGMV} = \log(\text{GMV}+1)$) was evaluated via Box-Cox analysis~\cite{Box1964} and confirmed as the optimal normalizing transformation for our data. However, it changes the estimand to $\mathbb{E}[\log(Y+1)]$, can disagree directionally with raw GMV when the variance of log-outcomes shifts between treatment and control, and disproportionately captures tail behavior at the expense of low-gifting user dynamics. Trimmed means change the estimand to a conditional mean. Non-parametric tests improved stability but reduced interpretability. Post-stratification preserves the original metric scale and population estimand while delivering substantially larger power gains.

\section{Practical Lessons, Deployment Guidance and Limitations}
\label{sec:deployment}

\textbf{Implementation:} Deploying post-stratification requires a pre-period data pipeline and automated jobs to calculate population weights. We also implemented monitoring for "stratum drift" (significant changes in stratum sizes) and weight anomalies to prevent metric corruption.

\textbf{Applicability:} The method excels for ranking and recommendation experiments constrained by limited traffic, where heavy-tailed metrics otherwise require prohibitively long durations. It is most effective when outcome heterogeneity is predictable from history (e.g., monetization, gaming whales) but is \emph{not recommended} for experiments explicitly targeting tail users (e.g., VIP features), where the post-stratified estimand would systematically down-weight the very population under study.

\textbf{Guardrails:} We freeze stratification thresholds before analysis to prevent p-hacking and also enforce minimum stratum sizes (rejecting valid results if a stratum has $<50$ users). Both the post-stratified and raw metrics are surfaced to experiment owners to maintain interpretability and make tail-concentrated effects visible.

\section{Conclusion}
Reliable evaluation of ranking and recommendation systems becomes significantly more challenging when downstream monetization metrics exhibit heavy-tailed behavior. Standard A/B testing approaches and common variance-reduction techniques often lack sufficient sensitivity under realistic traffic constraints, slowing iteration on ranking-driven product improvements.

We presented a production-ready post-stratification framework that improves experimental sensitivity while preserving business interpretability. Deployed across large-scale IR experimentation workflows, the approach reduces variance and enables faster, more stable decisions without increasing traffic. Beyond statistical gains, our experience highlights the importance of practical guardrails, interpretable metrics, and seamless integration into existing experimentation pipelines.

This framework is broadly applicable to IR systems where outcomes are driven by rare but high-impact user behavior, including engagement metrics like long-tail query engagement, marketplace liquidity measures, gaming metrics, and ads conversion value. The method is most effective when outcome heterogeneity is predictable from pre-treatment covariates. 


\section*{Acknowledgments}
The authors acknowledge the peoples of the Woi Wurrung and Boon Wurrung language groups of the eastern Kulin Nation on whose unceded lands ACM SIGIR 2026 was hosted. We pay our respects to their Elders past and present, and extend that respect to all Aboriginal and Torres Strait Islander peoples today and their continuing connection to land, sea, sky, and community.

\section*{About the presenter}
Neeti Pokharna is a Staff Data Scientist at ShareChat, specializing in causal inference and experimentation for large-scale recommender systems. She leads the development of metric sensitivity frameworks and evaluation methodologies that drive product decision-making. An alumna of Birla Institute of Technology and Science Pilani, Indian Statistical Institute Kolkata, Indian Institute of Technology Kharagpur, and Indian Institute of Management Calcutta, Neeti is an active contributor to the data science community, sharing her learnings through talks at various universities across India.
\bibliographystyle{ACM-Reference-Format}
\bibliography{references}
\end{document}